\documentclass{article} 
\usepackage{iclr2021_conference,times}

\usepackage{amsmath,amsfonts,bm}

\def\eqref#1{equation~\ref{#1}}

\def\1{\bm{1}}

\DeclareMathAlphabet{\mathsfit}{\encodingdefault}{\sfdefault}{m}{sl}
\SetMathAlphabet{\mathsfit}{bold}{\encodingdefault}{\sfdefault}{bx}{n}

\usepackage{hyperref}
\usepackage{url}
\usepackage{graphicx}
\usepackage{caption}
\usepackage{subcaption}
\usepackage{soul}
\title{Comparing Visual Reasoning in Humans and AI}

\author{Shravan Murlidaran \\
Psychological \& Brain Sciences\\
University of California\\
Santa Barabra,  USA \\
\texttt{smurlidaran@ucsb.edu} \\
\And
William Yang Wang \\
Computer Science \\
University of California\\
Santa Barbara, USA \\
\texttt{william@cs.ucsb.edu}
\And
Miguel P. Eckstein  \\
Psychological \& Brain Sciences\\
University of California\\
Santa Barabra,  USA \\
\texttt{miguel.eckstein@psych.ucsb.edu} \\

}

\iclrfinalcopy 
\begin{document}

\maketitle
\begin{abstract}
Recent advances in natural language processing and computer vision have led to AI models that interpret simple scenes at human levels. Yet, we do not have a complete understanding of how humans and AI models differ in their interpretation of more complex scenes. We created a dataset of complex scenes that contained human behaviors and social interactions. AI and humans had to describe the scenes with a sentence.  We used a quantitative metric of similarity between scene descriptions of the AI/human and ground truth of five other human descriptions of each scene. Results show that the machine/human agreement scene descriptions are much lower than human/human agreement for our complex scenes. Using an experimental manipulation that occludes different spatial regions of the scenes, we assessed how machines and humans vary in utilizing regions of images to understand the scenes. Together, our results are a first step toward understanding how machines fall short of human visual reasoning with complex scenes depicting human behaviors.
\end{abstract}

\section{Introduction}
Over the last decade, visual reasoning by AI has seen major strides of advancements with the help of huge datasets of image-text pairs \citep{lin_microsoft_2015, sharma_conceptual_2018, agrawal_vqa_2016, zellers_recognition_2019}. These datasets are used to train AI models on image captioning, visual question answering, and image text retrieval. Especially in the last couple of years, the usage of Transformer architectures \citep{vaswani_attention_2017} has led to the rise of task agnostic models \citep{lu_vilbert_2019,  chen_uniter_2020, li_oscar_2020} that can encode joint vision-language features which can be fine-tuned to any downstream tasks to reach state-of-the-art performances. 

Despite these models' tremendous achievements, little is known about how it functions or how it compares to human visual reasoning. Understanding the mechanisms of these models is important to predict their behavior in different scenarios. For most of the AI literature, humans and AI models have been compared with each other only based on their performances on a given dataset. But using human performance as a criterion to judge these models' intelligence is not sufficient, as shown by many adversarial examples \citep{balda_adversarial_2020} where models that have above human performance fail in non-human-like ways. 

Recent studies have compared humans and deep neural networks across a variety of tasks, including contour detection, visual reasoning with synthetic tasks \citep{firestone_performance_2020}, abstraction and reasoning \citep{chollet_measure_2019}, visual search in real-world scenes \citep{eckstein_humans_2017}, areal scenes \citep{deza_assessment_2019} and medical images \citep{lago_under-exploration_2021}. Other studies have developed ways to get human attention maps by allowing humans to sharpen regions in blurred images \citep{das_human_2016} or click on important regions \citep{linsley_what_2017}. These maps were used as a metric to compare the attention maps generated by vision + language models. In the current study, we first selected a data set requiring more complex reasoning of human behaviors and social interactions than the commonly used MS COCO data set \citep{lin_microsoft_2015}. Second, we developed a paradigm for which we mask (i.e., occlude) regions of the scene and assess the impact on their descriptions by humans and a state-of-the-art image captioning model, OSCAR \citep{li_oscar_2020}. We propose this manipulation as a counterfactual occlusion \citep{fu_counterfactual_2020} that can reveal important differences between human and AI visual reasoning and could potentially be used for data augmentation and improved AI reasoning.

\section{Methods}
\begin{figure}[!hbt]
    \centering
    \includegraphics[width=\textwidth]{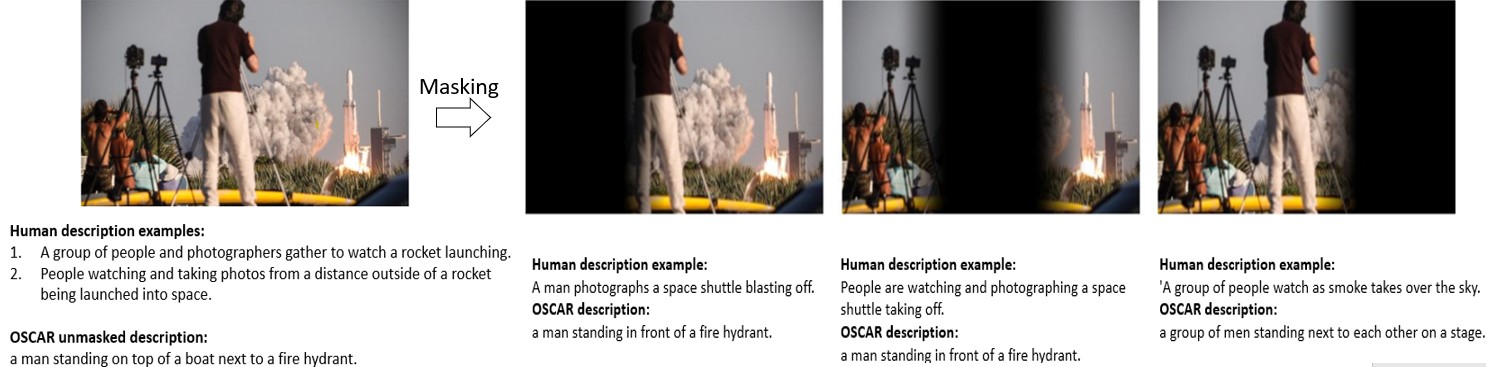}
    \caption{Example of descriptions from Humans and OSCAR in the masked and unmasked conditions}
    \label{fig:Dataset}
\end{figure}

\subsection{Dataset}
In this study, we used a scene description task to understand visual reasoning in both humans and AI models. The scenes/images used in the description task were collected by ten research assistants (RA). The RAs were instructed to choose image frames from movie scenes from which they can ask a question that will require complex reasoning of social interaction. They were given constant feedback on their submitted images to ensure that the scenes are indeed complex. A total of thousand images were collected, of which 500 randomly selected images were used in this study. An additional five RAs provided descriptions to these 500 images that were used as the ground truth. The authors inspected the descriptions to ensure students followed the instructions correctly.

\subsection{OSCAR}
OSCAR \citep{li_oscar_2020} is a Visual-Language model that is pre-trained on 6.5 million text-image pairs. It is one of the state-of-the-art models for scene description that uses co-attentional transformers to produce a combined feature space of image and text. This combined feature space is later fine-tuned to generate captions for a given image. In this study, we are using an OSCAR model already fine-tuned for scene description tasks. We use OSCAR to generate descriptions for the images in our dataset.

\subsection{Image perturbations}
We hand-picked a subset of forty-six images from our dataset for this part of the study. We qualitatively ensured that these images depicted human behaviors and/or social interactions that required complex reasoning to provide a description. To assess the contributions of different regions of the scene to a description, we masked different areas of the image. We analyzed how the human or OSCAR's description of the scene changed. Masks consisted of a black opaque spatial window with a soft edge that covered ~ 1/3 of the image and varied in its position (Fig.\ref{fig:Dataset}). These masked images were given as input to OSCAR to generate captions.

\subsubsection{Human descriptions to masked images}
We ran an Amazon Mechanical Turk study on these masked images to get descriptions from 45 people living in the United States. To minimize possible memory effects, each person viewed each image just once with the mask covering one of the three locations randomly. On average, there were fifteen observers per mask location for each image. 

\section{Results}

\begin{figure}[!hbt]
    \centering
     \begin{subfigure}[b]{0.49\textwidth}
         \centering
         \includegraphics[width=\textwidth]{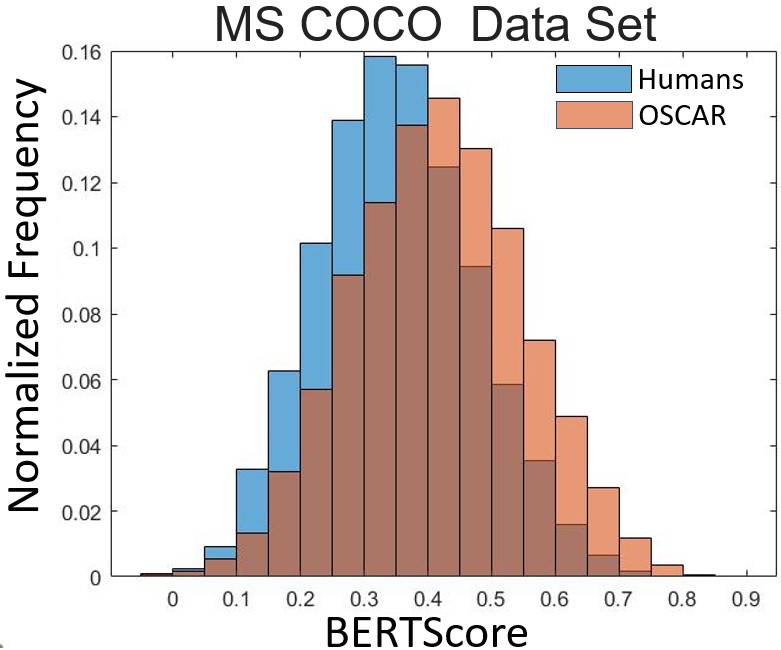}
         \caption{}
         \label{fig:Ours}
     \end{subfigure}
     \hfill
    \begin{subfigure}[b]{0.49\textwidth}
         \centering
         \includegraphics[width=\textwidth]{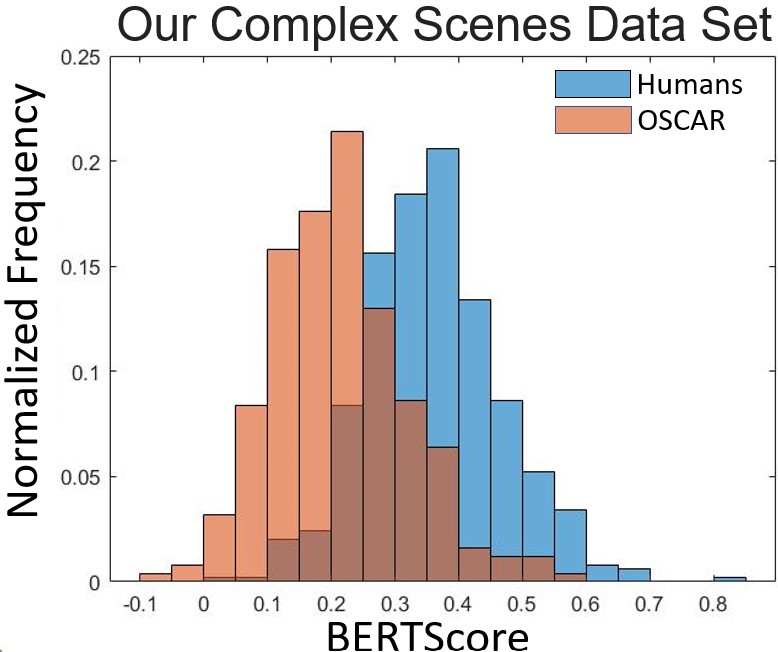}
         \caption{}
         \label{fig:MSCOCO}
    \end{subfigure}
    \caption{Human and OSCAR description similarity score (BERTScore) with five reference description for MS COCO and the UCSB complex scene data set.}
    \label{fig:DatasetValidation}
\end{figure}
\subsection{Comparing Descriptions}
We evaluated the similarity of descriptions in our experiments using the BERTScore \citep{zhang_bertscore_2020}. The BERTScore computes the cosine similarity of each word embedding from one sentence (a reference) to each word embedding in another sentence (a candidate). This word embedding is a contextual feature representation that constitutes the BERT architecture's output (Bidirectional Encoder Representations from Transformers). In the upcoming sections, we will be using BERTScore (BTS) as a performance metric to compare descriptions produced by humans and OSCAR across different conditions. A relatively higher BERTScore between a reference and candidate sentence indicates that they are more similar in meaning and, therefore, better performance. BERTScore can take values from -1 to 1. To ensure interpretability, the BERTScore was re-scaled to the interval [0,1] using a base value which is computed by finding the average BERTScore of randomly paired sentences.
\subsection{Visual reasoning complexity of present dataset vs. MS COCO}
The MS COCO dataset \citep{lin_microsoft_2015} is one of the largest publicly available datasets in the AI literature. It consists of over 300K labeled images of scenes containing common objects in their natural context and five human descriptions used as ground truth. 

To assess our dataset's visual reasoning complexity, we compared the performance of humans and OSCAR in both datasets. For humans, we used the five descriptions for each image in both datasets and computed a one vs. all BERTScore for each human, and averaged the score across all ten possible comparisons. For OSCAR, we obtained the BERTScore of its description with the corresponding human description in each dataset and averaged across the comparisons to the five humans. Fig.\ref{fig:DatasetValidation} shows that the similarity in the descriptions are almost the same between human pairs in MS COCO dataset(M=0.35, SD=0.12) and our dataset(M=0.36,SD=0.11), while OSCAR is considerably less similar to humans in our dataset(M=0.21, SD=0.11) than MS COCO(M=0.42, SD=0.13).
\begin{figure}[!hbt]
    \centering
    \includegraphics[width=\textwidth]{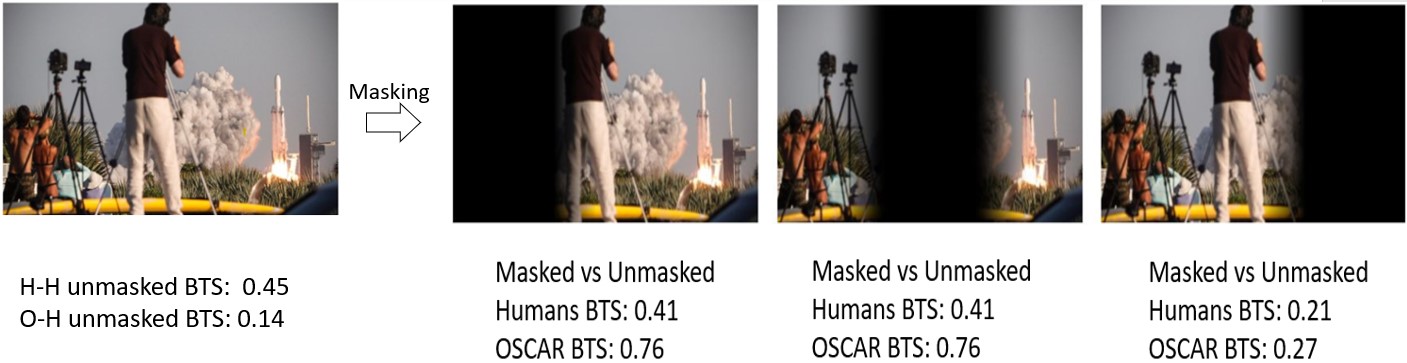}
    \caption{a) Example of description similarity score (BERT Score, BTS) for Oscar and humans relative to five human reference descriptions, b) Example of description similarity score (BERT Score, BTS) for masked images relative to the unmasked image reference descriptions. Human BERT Scores for masked images are computed relative to human unmasked image descriptions. Oscar BERT Scores for masked images are computed relative to OSCAR's description to the unmasked images.
    \label{fig:MaskBERTScores}
}
\end{figure}
\subsection{How do humans and OSCAR's masked performance compare?}
To assess the masks' effect on visual reasoning, we used the mean and standard deviation of the BERTScores across mask positions. For humans, each image's mask position had approximately 15 descriptions which were compared to the original five unmasked descriptions (a total of 75 BERTScores per mask position for each image). For each image and mask position, we computed average BERTScores across the 75 scores. For OSCAR, each mask position in each image produced one description, which was compared with the unmasked OSCAR description. Therefore, we had a total of six BERTScores for each image (three humans and three OSCAR corresponding to the three masked conditions).  Fig.\ref{fig:MaskBERTScores} shows an example of the resulting BERTScores (BTS) for the three mask conditions for humans and OSCAR. 

\begin{figure}[!hbt]
    \centering
     \begin{subfigure}[b]{0.42\textwidth}
         \centering
         \includegraphics[width=\textwidth]{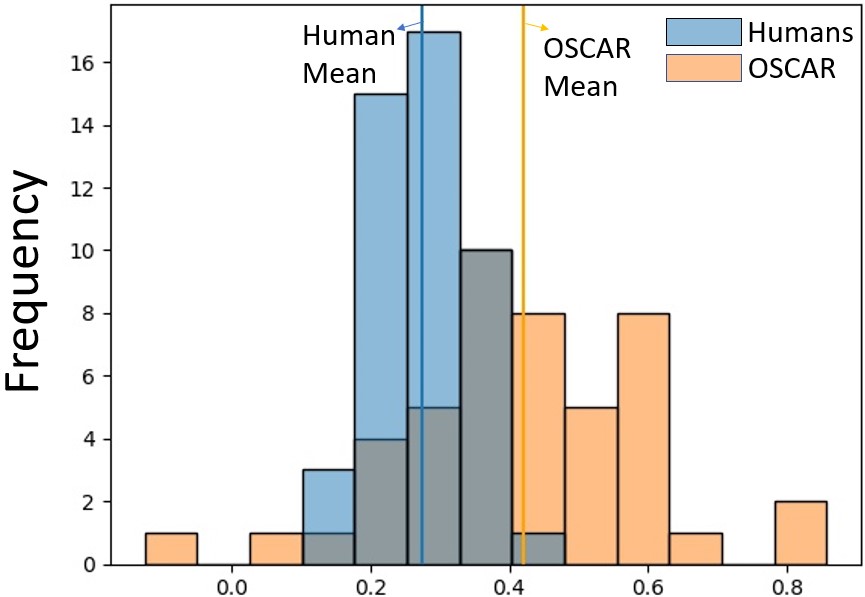}
         \captionsetup{justification=centering}
         \caption{Mean BERT Score across mask positions.}
         \label{fig:Mean}
     \end{subfigure}
     \hfill
    \begin{subfigure}[b]{0.42\textwidth}
         \centering
         \includegraphics[width=\textwidth]{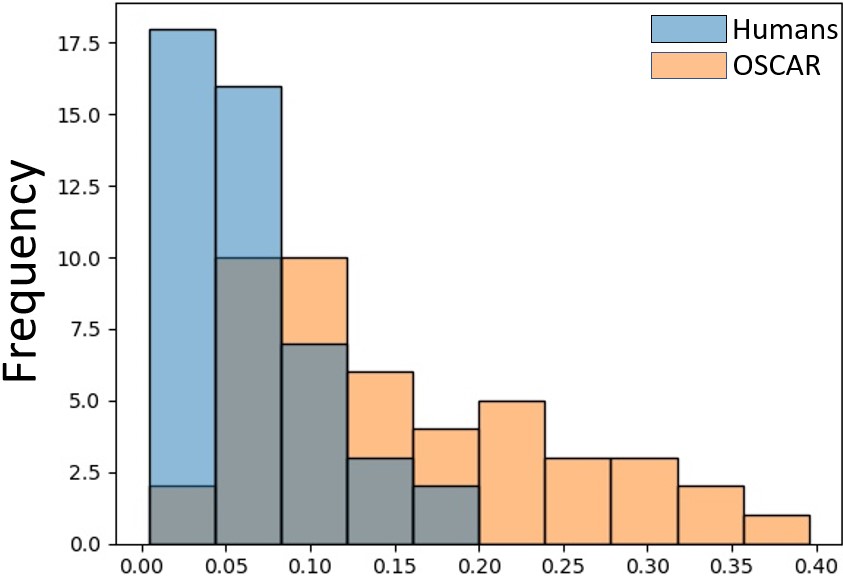}
         \captionsetup{justification=centering}
         \caption{Standard Deviation of BERTScore across mask positions.}
         \label{fig:SD}
    \end{subfigure}
    \caption{Mean (a) and Standard Deviation (b) across mask positions of description similarity scores  BERT Score (relative to unmasked image descriptions).}
    \label{fig:MaskMeanAndSD}
\end{figure}

Fig.\ref{fig:MaskMeanAndSD} shows the histogram of the mean and standard deviation of BERTScores across mask positions.
Fig.\ref{fig:Mean} shows that on average, OSCAR's scores are less impacted by the presence of the masks than humans (mean human BTS = 0.27  vs. mean OSCAR BTS = 0.42). In addition the effect on OSCAR's scores varies greatly across images (sdev = 0.16) while that on human descriptions is more consistent across images (sdev = 0.06).
Fig.\ref{fig:SD} shows that OSCAR's descriptions across mask positions vary more than humans. This might suggest that humans can extract scene descriptions from various regions/objects from a scene, while OSCAR might rely on more localized objects/regions.

\begin{figure}[!hbt]
    \centering
     \begin{subfigure}[b]{0.5\textwidth}
         \centering
         \includegraphics[width=\textwidth]{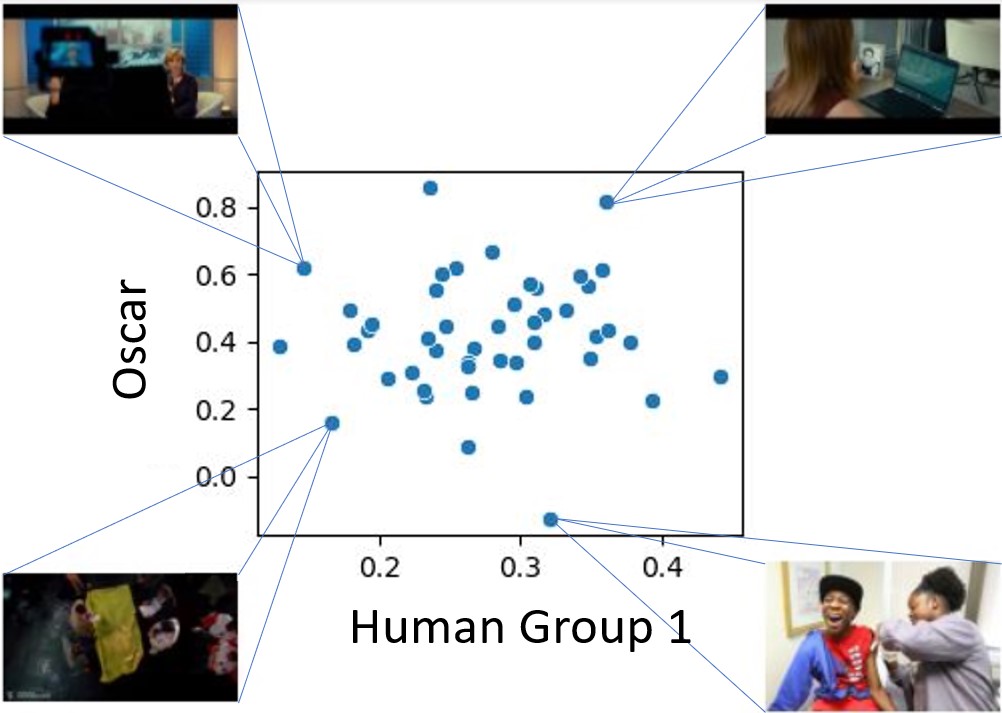}
         \caption{}
         \label{fig:HVsO}
     \end{subfigure}
     \hfill
    \begin{subfigure}[b]{0.49\textwidth}
         \centering
         \includegraphics[width=\textwidth]{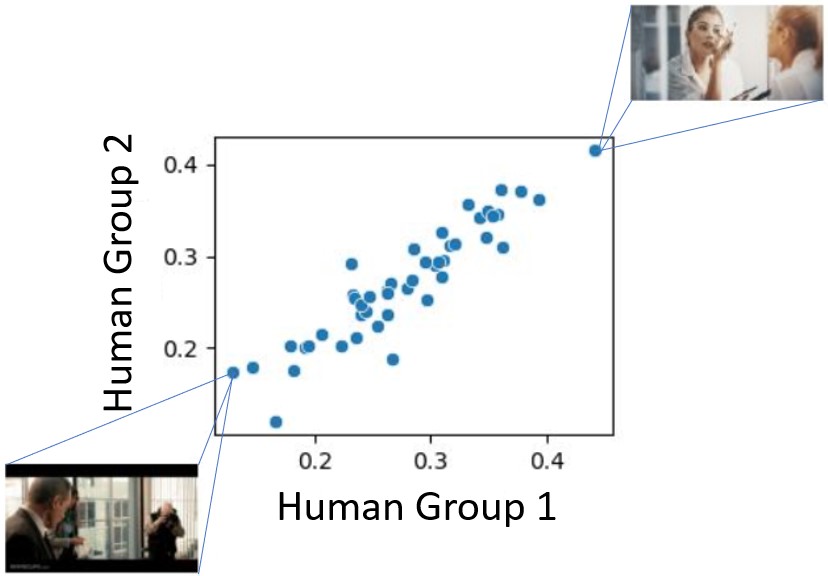}
         \caption{}
         \label{fig:HVsH}
    \end{subfigure}
    \caption{a) Average effect of the masks (across positions) on description similarity for each image for OSCAR vs. a sample human group 1 (averaged across observers). b)  Average effect of the masks (across positions) on description similarity for each image for two sample groups of humans (averaged across observers).}
    \label{fig:PerfCorrPlots}
\end{figure}

In order to compare the average effect of masks on humans and OSCAR, we divided the humans into two equal groups and correlated the average score across mask conditions for one human group with the average score across mask conditions for OSCAR and the other different human group. We repeated this analysis for a total of 10000 unique group splits for humans and found the required correlations. We also used bootstrap re-sampling of images (10,000 samples) for each split to account for variability across images. Fig.\ref{fig:PerfCorrPlots} shows an example scatter plot of average scores across mask positions for a sample split of the human into two groups and for OSCAR vs human group 1. The results show that the average effect of the mask on the descriptions for each image is significantly correlated between humans (r = 0.87, p \textless0.001) while there is no significant correlation with OSCAR (r = 0.045, p = 0.39).

\section{Conclusion}
Although our results confirm that a state-of-the-art image captioning model (OSCAR) performs similarly to humans for the MS COCO data, the findings show that the model falls short of humans for images depicting human behavior and social interactions that require more complex visual reasoning. Our masking manipulation suggests fundamental differences between human and OSCAR's visual commonsense reasoning and suggests a higher integration of information across the scene for humans. We suggest that training models to match human reasoning across masking conditions (a form of counterfactual reasoning) might improve overall AI visual reasoning capabilities.

\section{Acknowledgements}
The research was funded by the U.S. Army Research Office under Contract Number W911NF-19-D-0001 for the Institute for Collaborative Biotechnologies.  The views and conclusions contained in this document are those of the authors and should not be interpreted as representing the official policies, either expressed or implied, of the U.S. Government. The U.S. Government is authorized to reproduce and distribute reprints for Government purposes notwithstanding any copyright notation herein.

\bibliography{ICLR}
\bibliographystyle{iclr2021_conference}

\end{document}